\documentclass[conference]{IEEEtran}
\IEEEoverridecommandlockouts
\usepackage{cite}
\usepackage{amsmath,amssymb,amsfonts}
\usepackage{graphicx}
\usepackage{textcomp}
\usepackage{booktabs}
\usepackage{xcolor}
\usepackage{eso-pic}

\def\BibTeX{{\rm B\kern-.05em{\sc i\kern-.025em b}\kern-.08em
    T\kern-.1667em\lower.7ex\hbox{E}\kern-.125emX}}
\begin{document}
\bstctlcite{IEEEexample:BSTcontrol}

\title{Conflict-Aware Retrieval-Augmented Generation: Knowledge Conflict Detection, Reranking, and Adaptive Context Selection}

\author{\IEEEauthorblockN{Alaa Zahweh}
\IEEEauthorblockA{\textit{Department of Science} \\
\textit{Lebanese University}\\
Beirut, Lebanon \\
alaa.zahwi@st.ul.edu.lb}
\and
\IEEEauthorblockN{Hadi Hasan}
\IEEEauthorblockA{\textit{Department of Electrical and} \\
\textit{Computer Engineering} \\
\textit{American University of Beirut}\\
Beirut, Lebanon \\
hsh24@mail.aub.edu}
\and
\IEEEauthorblockN{Ali Chehab}
\IEEEauthorblockA{\textit{Department of Electrical and} \\
\textit{Computer Engineering} \\
\textit{American University of Beirut}\\
Beirut, Lebanon \\
chehab@aub.edu.lb}
}

\maketitle

\begin{abstract}
Retrieval-Augmented Generation (RAG) systems retrieve document chunks by semantic similarity alone, with no mechanism to detect when retrieved chunks contradict each other. We extend a conventional PDF-based RAG system with conflict detection at two pipeline stages: at ingestion time, where a new chunk is compared against similar existing chunks by an LLM-as-a-judge and detected conflicts are resolved by the user before the chunk is admitted, and at retrieval time, where conflicts among top-ranked chunks are surfaced as warnings alongside the answer. We also implement two reranking strategies, a point-wise cross-encoder and a list-wise LLM reranker, and a score-cliff rule that sets context length from cross-encoder scores. The ingestion-time judge reaches an F1 of 81.2\% on FEVER and 95.0\% on SICK sentence pairs. On multi-claim passages its recall holds near 85\% while its precision falls from 92\% to 82\%, on PDF chunk pairs it attains 95.1\% precision and 78.7\% recall, and it detects 67.4\% of real Wikipedia contradictions. Under the standard TREC-DL protocol, with BM25 retrieval over the full MS MARCO collection, both rerankers raise nDCG@10 from 49.1 to 70.6 and are statistically indistinguishable. The score-cliff rule improves context F1 from 0.57 to 0.64 only after recalibration. End to end, on documents with contradictory revisions, the ingestion check cuts contaminated contexts from 85\% to 18\% of the affected questions, the list-wise reranker outperforms the cross-encoder by 11.6 points of answer accuracy, and the truncation gain does not carry over to answers.
\end{abstract}


\section{Introduction}
Retrieval-Augmented Generation (RAG) has become a standard architecture for grounding large language model (LLM) outputs in external knowledge, particularly in document question-answering systems where users upload their own documents and query them directly. Two properties of the pipeline determine answer quality: whether the retrieved chunks are mutually consistent~\cite{xie2024adaptive,jin2024tug}, and whether the most relevant chunks surface~\cite{sun2023chatgpt}. When either property fails silently, the output can look fluent and well cited while remaining factually unreliable, a concern that is sharpest in high-stakes domains such as healthcare~\cite{hasan2026toward}.

Consider an organization whose document repository is updated over time, for example policy documents, financial reports, or product specifications revised across fiscal years. A user who uploads a new specification may unknowingly introduce information that contradicts a document uploaded months earlier. A standard RAG system has no way to detect this: both chunks are semantically similar to a future query, both may be retrieved, and the LLM must decide, with no explicit signal, which chunk to trust.

In this work we extend a conventional PDF-based RAG system along both dimensions. We add conflict detection at ingestion time, where a new chunk is compared against similar existing chunks by an LLM-as-a-judge and detected conflicts are resolved by the user, and at retrieval time, where conflicts among the top reranked chunks are flagged as warnings. Between retrieval and generation we implement two reranking strategies, a point-wise cross-encoder and a list-wise LLM reranker, and let the reranker scores set the context length.

We evaluate the system at the level of its components and as a whole: the conflict detector on sentence pairs, multi-claim passages, real Wikipedia contradictions, and PDF documents passing through the ingestion path; the rerankers under the standard TREC-DL protocol over the full MS MARCO collection; the truncation rule against fixed top-$k$; and the assembled pipeline on question answering over a collection into which contradictory document revisions are uploaded. Our contributions are (i) a two-stage conflict-detection pipeline with human-in-the-loop resolution in place of automated adjudication; (ii) a characterization of an LLM judge from sentence pairs to PDF chunks that shows where its precision and recall are lost; (iii) a comparison of point-wise and list-wise reranking under the standard protocol, with measured latency, cost, and failure rates, and an empirical evaluation of score-cliff truncation that corrects its original setting; and (iv) an end-to-end evaluation that separates what carries over to answers from what does not.

\section{Related Work}
\label{sec:related}

\textit{Knowledge conflicts.} Xie et al.~\cite{xie2024adaptive} showed that coherent counter-evidence makes models highly receptive to incorrect external content. Jin et al.~\cite{jin2024tug} and ConflictBank~\cite{su2024conflictbank} report availability and confirmation biases and a greater susceptibility of large models to conflicting evidence, and ClashEval~\cite{wu2024clasheval} shows that subtle numerical perturbations mislead models far more reliably than implausible ones, much as small, hard-to-detect perturbations mislead deep networks in adversarial settings~\cite{hasan2026defending}. An explicit conflict-detection step is therefore preferable to relying on the generating model to notice the disagreement. Resolution methods include Astute RAG~\cite{wang2025astute}, which consolidates internal knowledge with external passages; TruthfulRAG~\cite{liu2026truthfulrag} and FaithfulRAG~\cite{zhang2025faithfulrag}, which model conflicts at the level of facts; cross-validation reranking~\cite{wu2025eliminating}; and BordaRAG~\cite{li2025bordarag}, which treats aggregation as voting. In all of them the model resolves the conflict automatically at generation time, whereas we leave the resolution to a person. SummaC~\cite{laban2022summac} establishes that NLI-based inconsistency detection is far more effective at sentence-level than at document-level granularity, which motivates our pairwise, chunk-level detector and makes the granularity at which a detector is evaluated a question in its own right; entity substitution~\cite{longpre2021entity} and WikiContradict~\cite{hou2024wikicontradict} provide passage-level conflict data for that purpose.

\textit{Reranking.} The supervised multi-stage baseline was established by monoBERT/duoBERT~\cite{nogueira2019multi}. RankGPT~\cite{sun2023chatgpt} introduced list-wise permutation generation with a sliding window, which our list-wise reranker follows; PRP~\cite{qin2024large} reduces ranking to pairwise comparisons, set-wise methods~\cite{zhuang2024setwise} compare small candidate sets, FIRST~\cite{chen2025accelerating} ranks from first-token logits, and RankZephyr~\cite{pradeep2023rankzephyr} distills list-wise ranking into a small open model. GenRT~\cite{xu2024list} argues that reranking and truncation should be modeled jointly, and an applied study in financial QA~\cite{cheng2026enhancing} reports that cross-encoder reranking improves answer correctness. Conflict-resolution work assumes retrieval has already occurred, while reranking work is evaluated purely on relevance; we place both concerns in one pipeline and evaluate them together.

\begin{figure}[t]
    \centering
    \includegraphics[width=0.94\columnwidth]{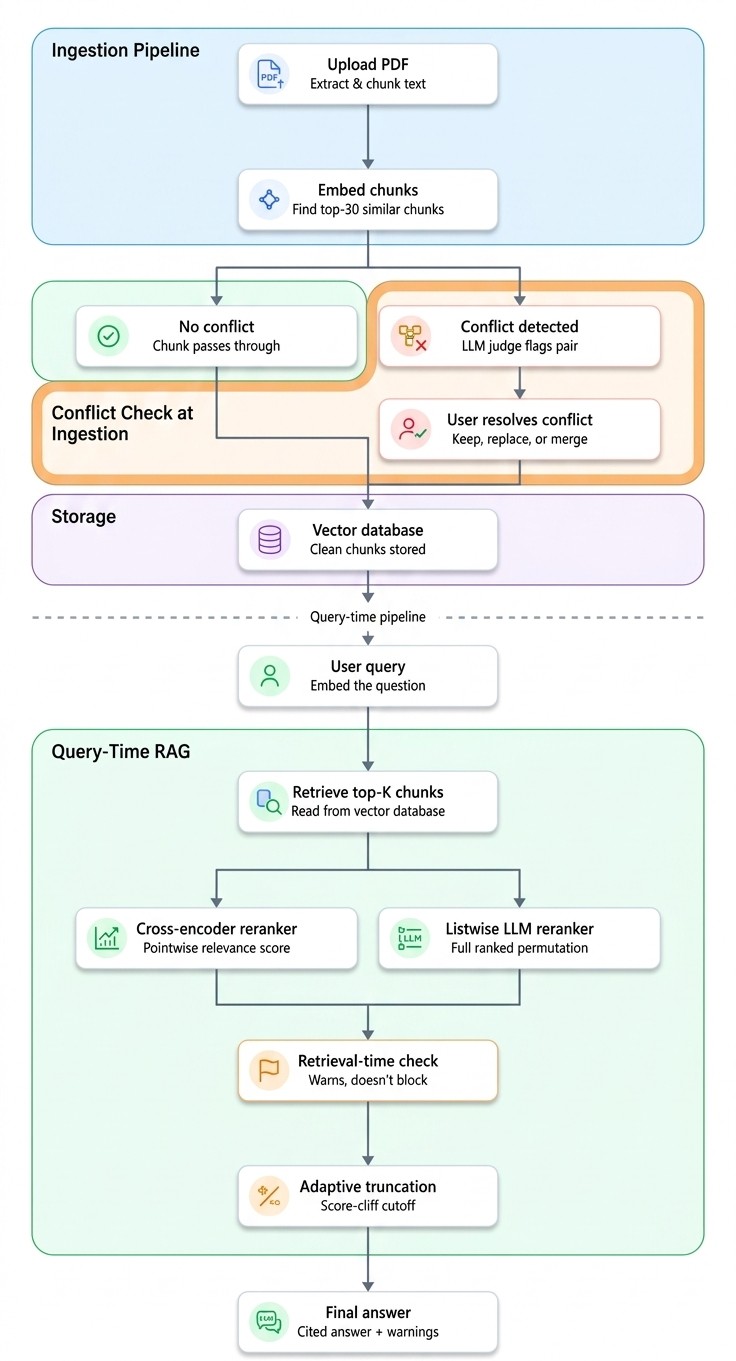}
    \caption{The conflict-aware RAG pipeline.}
    \label{fig:architecture}
\end{figure}

\section{Methodology}
\label{sec:methodology}

The system extends a conventional PDF-based RAG application with conflict detection at two pipeline stages and a configurable reranking stage (Fig.~\ref{fig:architecture}). The two conflict-detection stages are intentionally asymmetric. The ingestion-time stage is \emph{blocking}: a conflicting chunk has not yet entered the knowledge base and can still be withheld. The retrieval-time stage is \emph{advisory}: an answer is already being composed, and flagging the disagreement is less disruptive than halting generation.

\subsection{Base Pipeline}

Uploaded PDFs are parsed with PyMuPDF, split into sentence-aligned chunks of at most 1{,}000 characters, embedded with \texttt{text-embedding-3-small}, and stored in ChromaDB (cosine similarity) with source and page metadata. Database writes, including conflict checks, are serialized through a single worker, so that two concurrent uploads cannot each pass the check against a snapshot that omits the other. At query time a candidate pool of 20 chunks is retrieved, reranked, cut to a fixed top-5 or truncated adaptively, and passed to \texttt{gpt-4o-mini} to generate the answer.

\subsection{Ingestion-Time Conflict Detection and Resolution}
\label{sec:ingest}

Before a new chunk is written to the database, its top-30 nearest neighbors are retrieved and filtered to cross-file candidates above a cosine similarity of $\tau_s = 0.55$; same-file candidates are excluded because overlapping chunks within one document are expected. Survivors are passed pairwise to an LLM-as-a-judge (\texttt{gpt-4o-mini}, temperature 0, JSON output) that returns a verdict, a confidence, and a rationale. The prompt asks the judge to identify claims about the same entity, metric, and time period before assessing disagreement, and includes eight few-shot examples covering direct contradictions, complementary details, and ambiguous entity-identity cases. A conflict verdict is accepted above a confidence threshold of $\tau_c = 0.40$, set deliberately low: because a human resolves every flagged conflict, a false positive costs one extra decision while a false negative silently admits a real contradiction, an asymmetry that is most acute where AI systems support clinical decisions~\cite{hasan2026end,hasan2026comprehensive}. Section~\ref{sec:results} tests whether this threshold controls the judge's operating point.

When conflicts are accepted, the chunk is withheld and all of its conflicts are grouped into a single review item showing both texts and the judge's reasoning. The user may keep the existing chunk; keep the new chunk, removing the entry it replaces; keep both, tagging them with a shared conflict-group id so the disagreement is recognized at retrieval time as deliberately retained; or supply a custom replacement. Every detection and resolution event is appended to an audit log.

\subsection{Reranking Strategies}

\textit{Point-wise cross-encoder:} each candidate chunk is scored jointly with the query by \texttt{bge-reranker-base}~\cite{xiao2024cpack} in one batched forward pass; the score is retained with each chunk and used by adaptive truncation. \textit{List-wise LLM reranking:} following RankGPT~\cite{sun2023chatgpt}, all candidates are shown to \texttt{gpt-4o-mini} with numeric identifiers, and the model outputs a ranked permutation. Any identifier the model omits is appended in its original retrieval order, so a malformed response never fails the pipeline. The deployed pool of 20 fits in one call; larger pools use the sliding window of~\cite{sun2023chatgpt} (window 20, step 10).

\subsection{Retrieval-Time Conflict Detection}

After reranking, the top five chunks are checked pairwise, skipping same-file pairs, by a second judge (four few-shot examples) that requires four conditions simultaneously: the same entity, the same time period, the same attribute, and mutually exclusive values. It is intended to be stricter than the ingestion-time judge, an intent that Section~\ref{sec:results} examines. Pairs sharing a conflict-group id are reported as known conflicts without a judge call. Detected conflicts are attached to the answer as warnings and do not block generation.

\subsection{Adaptive Truncation}

The context length is set from the shape of the reranker's score distribution. Let $s_1 \geq \dots \geq s_n$ be the reranker scores of the candidate pool (the sigmoid of the cross-encoder logits), normalized to $\tilde{s}_i = (s_i - s_n)/(s_1 - s_n)$, with the relative drop between consecutive candidates defined as $\delta_i = 1 - \tilde{s}_{i+1}/\tilde{s}_i$. The system scans for the first \emph{score cliff}, a relative drop exceeding a threshold $\theta$ ($0.7$ in the original design), and truncates there, clipped to between 2 and 10 chunks:
\begin{equation}
k^\star = \mathrm{clip}\!\left(\min\{\, i : \delta_i > \theta \,\},\; 2,\; 10\right)\!.
\end{equation}
Because $\tilde{s}_n = 0$, the last gap always qualifies, so a pool with no earlier cliff yields the upper bound of ten. The rule applies to the cross-encoder configuration only: the list-wise reranker produces a rank position rather than a magnitude.

\section{Experimental Setup}
\label{sec:setup}

All experiments use a single implementation of the pipeline, with \texttt{gpt-4o-mini-2024-07-18} at temperature 0 and the cross-encoder on a laptop CPU (Intel Core Ultra 7 255U, no GPU). Both judge prompts were adjusted only on development samples of 150 pairs each from the FEVER and SICK training data, then frozen; every evaluation set was run once. We report 95\% bootstrap confidence intervals (CI) and two-sided paired randomization tests.

\textit{Conflict detection} is evaluated at four levels. (i)~Sentence pairs: 1{,}050 claim--evidence pairs from the \texttt{nli\_fever} dev split of FEVER~\cite{thorne2018fever,nie2019combining} (450 REFUTES, 450 SUPPORTS, 150 NOT ENOUGH INFO) and 1{,}050 pairs from the SICK test split~\cite{marelli2014sick} (450 contradiction, 450 entailment, 150 neutral); REFUTES and contradiction are conflicts. (ii)~Multi-claim passages (FEVER-MC): for 300 FEVER dev pages with at least one REFUTES and six SUPPORTS claims, the stored passage is the opening of the page's Wikipedia introduction (4.5 sentences on average) and the incoming passage concatenates $m$ human-written claims, either $m$ SUPPORTS claims or one REFUTES claim among $m{-}1$ SUPPORTS claims, for $m \in \{1,2,4,6\}$ with pages and stored passages held fixed (600 pairs per $m$). (iii)~Real contradictions: the 190 WikiContradict~\cite{hou2024wikicontradict} instances that involve two distinct Wikipedia passages (121 explicit, 69 requiring reasoning); all are conflicts, so only recall is defined. (iv)~PDF documents: from 500 NQ-Swap questions~\cite{longpre2021entity}, each with a passage whose answer entity is replaced by another of the same type, we assemble 100 five-passage documents, each in an original version and a revised version where two passages carry the substituted answer, and render all 200 as PDFs (the \emph{revision corpus}). They pass through the full ingestion path (860 chunks), one version of every document first and then the other, with the order of original and revised randomized per document. A chunk pair is a gold conflict when its chunks come from different versions of one document and both contain the answer mention of a substituted passage.

\textit{Reranking} follows the standard TREC-DL protocol~\cite{sun2023chatgpt}: for all 43 judged queries of TREC-DL 2019~\cite{craswell2020trec19} and all 54 of TREC-DL 2020~\cite{craswell2021trec20}, the top-100 passages retrieved by BM25 (Anserini, $k_1{=}0.9$, $b{=}0.4$) from the full 8.8M-passage MS MARCO collection~\cite{nguyen2016msmarco} are reranked, with unjudged passages counted as non-relevant. We report nDCG@10 and, at relevance grade $\geq 2$, MRR@10 and P@10, at depth 100 and at depth 20, the deployed pool size. Our BM25 baseline reproduces the published 50.58 and 47.96 nDCG@10~\cite{sun2023chatgpt} exactly.

\textit{Truncation} is evaluated on the same queries with the BM25 top-20 pool reranked by the cross-encoder, measuring the precision, recall, and F1 of the selected context against the relevant passages in the pool. The rule is compared with every fixed $k \leq 10$, with a per-query oracle, and with $k$ and $\theta$ selected on one track and applied to the other.

\textit{End to end}, the 500 questions of the revision corpus are asked against three knowledge bases: \emph{clean} (original versions only), \emph{unchecked} (all uploads admitted), and \emph{checked} (after the ingestion check, with a simulated user who resolves each true conflict in favor of the original version and marks each false alarm ``keep both''). Of the questions, 200 concern a substituted passage. An answer is correct if it contains the original answer and contaminated if it contains the substituted one.

\section{Results}
\label{sec:results}

\subsection{Conflict Detection}

\begin{table}[t]
\centering
\caption{Conflict detection (\%). WikiContradict contains only conflicts, so only recall is defined.}
\label{tab:conflict-results}
\begin{tabular}{l r ccc}
\toprule
Evaluation set & $n$ & P & R & F1 \\
\midrule
\multicolumn{5}{l}{\textit{Ingestion-time judge}} \\
FEVER, sentence pairs       & 1{,}050 & 79.4 & 83.1 & 81.2 \\
SICK, sentence pairs        & 1{,}050 & 95.5 & 94.4 & 95.0 \\
FEVER-MC, $m{=}6$           & 600     & 81.9 & 84.3 & 83.1 \\
PDF chunk pairs             & 617     & 95.1 & 78.7 & 86.1 \\
WikiContradict              & 190     & --   & 67.4 & --   \\
\midrule
\multicolumn{5}{l}{\textit{Retrieval-time judge}} \\
FEVER, sentence pairs       & 1{,}050 & 79.3 & 81.1 & 80.2 \\
SICK, sentence pairs        & 1{,}050 & 82.9 & 99.3 & 90.4 \\
FEVER-MC, $m{=}6$           & 600     & 80.4 & 52.0 & 63.2 \\
WikiContradict              & 190     & --   & 61.1 & --   \\
Top-5 chunk pairs (QA)      & 2{,}902 & 69.2 & 65.3 & 67.2 \\
\bottomrule
\end{tabular}
\end{table}

\begin{figure}[t]
    \centering
    \includegraphics[width=\columnwidth]{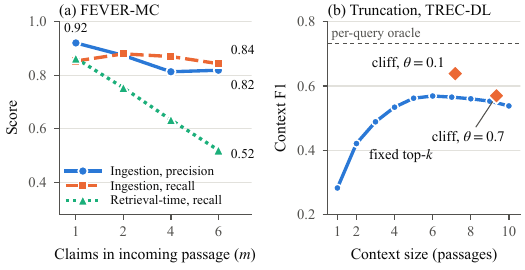}
    \caption{(a) Judge precision and recall as the incoming passage grows from one to six claims. (b) Context F1 of truncation policies on TREC-DL (BM25 top-20 pool, 97 queries).}
    \label{fig:panels}
\end{figure}

\textit{Sentence pairs.} The ingestion-time judge reaches an F1 of 81.2\% on FEVER (recall CI 79.5--86.5) and 95.0\% on SICK (Table~\ref{tab:conflict-results}). Its false positives concentrate in the ambiguous class: NOT ENOUGH INFO pairs are flagged 30.7\% of the time against 11.3\% for SUPPORTS pairs, and 19 of the 20 false positives on SICK are neutral pairs.

\textit{Multi-claim passages.} Recall does not degrade when the incoming passage carries several claims (Fig.~\ref{fig:panels}(a)): a contradiction buried among five consistent claims is found as often as an isolated one (84.3\% against 85.3\%). Precision falls from 92.1\% to 81.9\%, as false positives on consistent passages rise from 22 to 56 of 300. Multi-claim chunks therefore cost review effort rather than missed conflicts.

\textit{PDF documents.} In the ingestion path, 617 chunk pairs reach the judge (0.72 per chunk), which attains 95.1\% precision and 78.7\% recall on them. Of the 396 non-conflicting pairs, 333 are near-duplicates of unchanged text, 5 of them flagged, and 63 pair chunks of different documents, 4 of them flagged. The similarity filter loses one of the 211 conflicting chunks, so detection is bounded by the judge, and the 173 review items raised for 860 chunks include 7 false alarms. The judge's misses are typically chunk pairs identical apart from one name or number, a difference it notes and then dismisses as immaterial.

\textit{Real contradictions.} On WikiContradict the judge detects 67.4\% of the contradictions (75.2\% of the explicit ones, 53.6\% of those requiring reasoning), and only 84.2\% of these independently worded pairs exceed the similarity threshold, so filter and judge together would intercept 59.5\%.

\textit{Calibration.} Every conflict verdict on FEVER and SICK carries a self-reported confidence between 0.85 and 0.99, so the acceptance threshold of 0.40 never changes a verdict. The recall-leaning behavior is a property of the prompt alone.

\textit{Retrieval-time judge.} The conjunctive formulation does not make this judge stricter: it matches the ingestion-time judge on FEVER and is less precise on SICK, where it flags 49\% of neutral pairs. It is also fragile on multi-claim passages, its recall falling from 86.3\% to 52.0\% as $m$ grows from 1 to 6, because its four-condition format commits it to a single aligned claim pair.

\subsection{Reranking}

\begin{table}[t]
\centering
\caption{Reranking BM25 results on TREC-DL 2019/2020 (full collection, 97 queries). MRR@10 and P@10 are pooled; s/q is the median time per query, \$/1k the API cost per thousand queries.}
\label{tab:rerank-results}
\setlength{\tabcolsep}{3.3pt}
\begin{tabular}{l ccc cc rr}
\toprule
 & \multicolumn{3}{c}{nDCG@10} & & & & \\
\cmidrule(lr){2-4}
Run & '19 & '20 & All & MRR & P@10 & s/q & \$/1k \\
\midrule
BM25             & 50.58 & 47.96 & 49.12 & 67.5 & 37.7 & --   & --   \\
\midrule
\multicolumn{8}{l}{\textit{Top-20 reranked (deployed pool size)}} \\
Cross-encoder    & 64.47 & 60.64 & 62.34 & 84.0 & 47.1 & 3.7  & 0    \\
List-wise LLM    & 63.02 & 60.63 & 61.69 & 84.5 & 45.1 & 1.6  & 0.37 \\
\midrule
\multicolumn{8}{l}{\textit{Top-100 reranked}} \\
Cross-encoder    & 72.72 & 68.82 & 70.55 & 85.2 & 57.7 & 14.5 & 0    \\
List-wise LLM    & 73.58 & 68.21 & 70.59 & 84.2 & 55.8 & 15.1 & 3.31 \\
\bottomrule
\end{tabular}
\end{table}

Both rerankers improve significantly on BM25 at both depths ($p < 10^{-4}$, Holm-corrected; Table~\ref{tab:rerank-results}): reranking the top-100 raises pooled nDCG@10 from 49.12 to 70.55 with the cross-encoder and 70.59 with the list-wise reranker, within the range reported under this protocol~\cite{sun2023chatgpt}. The two are statistically indistinguishable at depth 100 ($p{=}0.98$) and at depth 20 ($p{=}0.47$). They differ operationally. At the deployed depth one list-wise call is faster than the cross-encoder on a laptop CPU (median 1.6\,s against 3.7\,s) but costs \$0.37 per thousand queries; at depth 100 it needs nine sequential calls. It is also less dependable: at depth 100, 27 of 873 window calls (3.1\%) returned no usable identifier. The cross-encoder has no API cost and always returns a complete ranking.

\subsection{Truncation}

At its original threshold ($\theta{=}0.7$) the score-cliff rule rarely acts (Fig.~\ref{fig:panels}(b)): for 84 of the 97 queries no cliff occurs among the first ten passages and the context is the upper bound of ten. Its context F1 of 0.569 is indistinguishable from a fixed top-5 (0.562, $p{=}0.71$) and equals the best fixed $k$ (0.569 at $k{=}6$). With $\theta$ selected on one track and applied to the other (0.1 in both directions), the rule keeps 7.2 passages on average and reaches 0.638, above the fixed $k$ selected the same way by 0.070 ($p < 0.001$), admitting 2.8 non-relevant passages per query instead of 4.7. The per-query oracle attains 0.732. We therefore revise the default to $\theta{=}0.1$ and carry both settings into the end-to-end experiment.

\subsection{End-to-End Question Answering}
\label{sec:e2e}

\begin{table}[t]
\centering
\caption{End-to-end question answering on the revision corpus (\%). Affected: the 200 questions about a substituted passage; Sub.: substituted chunk in the context; Cont.: answer contains the substituted value. Contexts are top-5 unless a cliff threshold is given.}
\label{tab:e2e}
\setlength{\tabcolsep}{3.4pt}
\begin{tabular}{ll ccc cc}
\toprule
 & & \multicolumn{3}{c}{Affected} & Unaff. & All \\
\cmidrule(lr){3-5}
KB & Ranking & Sub. & Acc. & Cont. & Acc. & Acc. \\
\midrule
Clean     & Dense                    & 0.0  & 66.5 & 0.0 & 64.3 & 65.2 \\
          & Cross-encoder (CE)       & 0.0  & 62.5 & 0.0 & 55.3 & 58.2 \\
\midrule
Unchecked & Dense                    & 85.0 & 61.0 & 3.5 & 64.0 & 62.8 \\
          & Cross-encoder (CE)       & 69.5 & 51.0 & 3.0 & 56.0 & 54.0 \\
          & List-wise LLM            & 92.0 & 61.5 & 3.0 & 68.3 & 65.6 \\
          & CE, cliff $\theta{=}0.7$ & 85.5 & 54.5 & 3.5 & 60.0 & 57.8 \\
          & CE, cliff $\theta{=}0.1$ & 72.0 & 49.0 & 3.5 & 54.7 & 52.4 \\
\midrule
Checked   & Dense                    & 18.0 & 63.5 & 0.5 & 63.0 & 63.2 \\
          & Cross-encoder (CE)       & 14.0 & 56.5 & 0.5 & 54.3 & 55.2 \\
\bottomrule
\end{tabular}
\end{table}

\textit{Ingestion-time check.} Admitting contradictory revisions degrades answers to the affected questions (Table~\ref{tab:e2e}): accuracy falls from 66.5\% to 61.0\% with the dense ranking ($p{=}0.054$) and from 62.5\% to 51.0\% with the cross-encoder ranking ($p < 0.001$). The generator seldom repeats the substituted value, consistent with implausible counter-evidence being easy to reject~\cite{wu2024clasheval}; it more often declines to answer, abstentions rising from 32.0\% to 42.5\% under the cross-encoder ranking ($p < 0.001$). After the ingestion check with simulated resolution, the substituted chunk reaches the context for 18.0\% of the affected questions instead of 85.0\%, and contaminated answers fall from 3.5\% to 0.5\% ($p{=}0.03$). Accuracy moves toward the clean level, by 2.5 points with the dense ranking and 5.5 with the cross-encoder, but with 200 questions neither gain is significant ($p{=}0.33$ and $p{=}0.051$).

\textit{Reranking.} The two rerankers, tied on TREC-DL, diverge here. The cross-encoder lowers accuracy relative to the dense retrieval order, by 7.0 points on the clean knowledge base and 8.8 on the unchecked one (both $p < 0.001$), whereas the list-wise reranker reaches 65.6\%, 11.6 points above the cross-encoder ($p < 0.001$) and 2.8 above the dense order ($p{=}0.10$). The answer-bearing chunk is among the top five for 89.0\% of questions in the dense order, 76.8\% after cross-encoder reranking, and 96.0\% after list-wise reranking. The cross-encoder's gain over BM25 thus does not transfer to a strong dense first stage over chunked documents. Chunks that straddle unrelated passages contribute, but the gap persists when the answer-bearing chunk is at least 80\% on the question's topic (in the top five for 82.0\% after cross-encoder reranking against 99.2\% in the dense order).

\textit{Truncation.} With the recalibrated threshold the context averages 5.4 chunks and accuracy does not differ from a fixed top-5 (52.4\% against 54.0\%, $p{=}0.42$): the context-F1 gain on TREC-DL does not turn into better answers here. The original threshold keeps 8.8 chunks and raises accuracy to 57.8\% ($p{=}0.03$) at 1.7 times the context tokens, a benefit of more context rather than of adaptive cutting.

\textit{Retrieval-time warnings.} On the unchecked knowledge base (the affected questions and 100 unaffected ones), a warning is raised for 108 of the 134 affected questions whose top five contain both versions of the asked fact (80.6\%), and for 11 of the 76 questions whose top five hold no conflicting pair (14.5\%), at a cost of 9.7 judge calls per query. Over the 2{,}902 chunk pairs judged, precision is 69.2\% and recall 65.3\% (Table~\ref{tab:conflict-results}).

\section{Discussion and Limitations}
\label{sec:discussion}

\textit{Conflict detection.} The ingestion-time judge is reliable when a contradiction is explicit, and loses recall when passages are worded independently or are identical apart from one entity; its false positives come from passages that share a topic without a comparable claim and multiply with the number of claims per chunk. Three limits remain. The PDF corpus tests single-entity substitutions in documents assembled from unrelated passages, not paraphrased revisions or topically coherent collections such as educational content repositories~\cite{hasan2025kag}. One judge model was used, so prompt and model effects are confounded. And because the acceptance threshold proved inert, the system has no working precision--recall control; a calibrated score or a second verification pass is needed to tune the review load. The retrieval-time judge offers no consistent precision advantage and loses recall on multi-claim chunks, so reusing the ingestion-time prompt at query time is the better design.

\textit{Reranking and truncation.} Component and end-to-end results disagree, and the latter are the more relevant to deployment: they favor the list-wise reranker for chunked documents, at the price of an API call and occasional malformed output. The reranking benchmark uses BM25 as first stage whereas the deployed system retrieves densely, and the end-to-end corpus is a single collection whose chunks mix topics. The recalibrated truncation threshold did not improve answers, so we claim no downstream benefit for it.

\textit{End-to-end evaluation.} The simulated user resolves every true conflict correctly, so the measured effect of the ingestion check is an upper bound on what a human reviewer would obtain. Substituted entities are often implausible, which lets the generator reject them and understates the harm of plausible contradictions, and answers are scored by string containment.

\section{Conclusion}
\label{sec:conclusion}

We presented a conflict-aware RAG system that adds contradiction detection at ingestion and retrieval time, human-in-the-loop resolution, two reranking strategies, and score-based truncation to a PDF-based pipeline, and evaluated it from its components to its answers. Several results hold up. An LLM judge at ingestion time detects contradictions between PDF chunks with 95.1\% precision and 78.7\% recall and, with simulated resolution, cuts contaminated contexts from 85\% to 18\% of the affected questions. Both rerankers add about 21 nDCG@10 points over BM25 under the standard TREC-DL protocol, and adaptive truncation, once recalibrated, exceeds every fixed context size in context F1. Other conclusions that component benchmarks alone would have suggested do not hold end to end. The cross-encoder ties with the list-wise reranker on TREC-DL but falls well behind it, and behind the unreranked dense order, on chunked documents. The truncation gain does not reach answer accuracy, and the accuracy gain from the ingestion check, although in the expected direction, is not statistically established. Future work includes a calibrated judge score, a retrieval-time check that reuses the ingestion-time prompt, rerankers validated against the deployed first stage, resolution studies with real users, and conflict corpora with paraphrased and topically coherent revisions.

\bibliographystyle{IEEEtran}
\bibliography{references}

@IEEEtranBSTCTL{IEEEexample:BSTcontrol,
  CTLuse_forced_etal = "yes",
  CTLmax_names_forced_etal = "6",
  CTLnames_show_etal = "1"
}

@inproceedings{xie2024adaptive,
  title={Adaptive chameleon or stubborn sloth: Revealing the behavior of large language models in knowledge conflicts},
  author={Xie, Jian and Zhang, Kai and Chen, Jiangjie and Lou, Renze and Su, Yu},
  booktitle={Proc. Int. Conf. Learn. Representations (ICLR)},
  year={2024}
}

@inproceedings{jin2024tug,
  title={Tug-of-war between knowledge: Exploring and resolving knowledge conflicts in retrieval-augmented language models},
  author={Jin, Zhuoran and Cao, Pengfei and Chen, Yubo and Liu, Kang and Jiang, Xiaojian and Xu, Jiexin and Qiuxia, Li and Zhao, Jun},
  booktitle={Proc. LREC-COLING},
  pages={16867--16878},
  year={2024}
}

@article{su2024conflictbank,
  title={{ConflictBank}: A benchmark for evaluating the influence of knowledge conflicts in {LLMs}},
  author={Su, Zhaochen and Zhang, Jun and Qu, Xiaoye and Zhu, Tong and Li, Yanshu and Sun, Jiashuo and Li, Juntao and Zhang, Min and Cheng, Yu},
  journal={arXiv:2408.12076},
  year={2024}
}

@inproceedings{wu2024clasheval,
  title={{ClashEval}: Quantifying the tug-of-war between an {LLM}'s internal prior and external evidence},
  author={Wu, Kevin and Wu, Eric and Zou, James},
  booktitle={Adv. Neural Inf. Process. Syst.},
  volume={37},
  pages={33402--33422},
  year={2024}
}

@inproceedings{wang2025astute,
  title={{Astute RAG}: Overcoming imperfect retrieval augmentation and knowledge conflicts for large language models},
  author={Wang, Fei and Wan, Xingchen and Sun, Ruoxi and Chen, Jiefeng and Arik, Sercan O},
  booktitle={Proc. 63rd Annu. Meeting Assoc. Comput. Linguistics (ACL)},
  pages={30553--30571},
  year={2025}
}

@inproceedings{liu2026truthfulrag,
  title={{TruthfulRAG}: Resolving factual-level conflicts in retrieval-augmented generation with knowledge graphs},
  author={Liu, Shuyi and Shang, Yu-Ming and Zhang, Xi},
  booktitle={Proc. AAAI Conf. Artif. Intell.},
  volume={40},
  number={38},
  pages={32168--32176},
  year={2026}
}

@inproceedings{zhang2025faithfulrag,
  title={{FaithfulRAG}: Fact-level conflict modeling for context-faithful retrieval-augmented generation},
  author={Zhang, Qinggang and Xiang, Zhishang and Xiao, Yilin and Wang, Le and Li, Junhui and Wang, Xinrun and Su, Jinsong},
  booktitle={Proc. 63rd Annu. Meeting Assoc. Comput. Linguistics (ACL)},
  pages={21863--21882},
  year={2025}
}

@inproceedings{wu2025eliminating,
  title={Eliminating retrieval knowledge conflicts: Cross-validation re-ranking with large language models},
  author={Wu, Qirui and Hai, Lin and Zheng, Hai-Tao and Xie, Ruobing and Yang, Saiyong and Sun, Xingwu and Kang, Zhanhui and Kim, Hong-Gee},
  booktitle={Proc. Int. Joint Conf. Neural Netw. (IJCNN)},
  pages={1--8},
  year={2025}
}

@inproceedings{li2025bordarag,
  title={{BordaRAG}: Resolving knowledge conflict in retrieval-augmented generation via {Borda} voting process},
  author={Li, Yuxin and Xu, Chen and Xu, Jun and Wen, Ji-Rong},
  booktitle={Proc. 34th ACM Int. Conf. Inf. Knowl. Manage. (CIKM)},
  pages={1737--1747},
  year={2025}
}

@article{laban2022summac,
  title={{SummaC}: Re-visiting {NLI}-based models for inconsistency detection in summarization},
  author={Laban, Philippe and Schnabel, Tobias and Bennett, Paul N and Hearst, Marti A},
  journal={Trans. Assoc. Comput. Linguistics},
  volume={10},
  pages={163--177},
  year={2022}
}

@article{nogueira2019multi,
  title={Multi-stage document ranking with {BERT}},
  author={Nogueira, Rodrigo and Yang, Wei and Cho, Kyunghyun and Lin, Jimmy},
  journal={arXiv:1910.14424},
  year={2019}
}

@inproceedings{sun2023chatgpt,
  title={Is {ChatGPT} good at search? {Investigating} large language models as re-ranking agents},
  author={Sun, Weiwei and Yan, Lingyong and Ma, Xinyu and Wang, Shuaiqiang and Ren, Pengjie and Chen, Zhumin and Yin, Dawei and Ren, Zhaochun},
  booktitle={Proc. Conf. Empirical Methods Natural Lang. Process. (EMNLP)},
  pages={14918--14937},
  year={2023}
}

@inproceedings{qin2024large,
  title={Large language models are effective text rankers with pairwise ranking prompting},
  author={Qin, Zhen and Jagerman, Rolf and Hui, Kai and Zhuang, Honglei and Wu, Junru and Yan, Le and Shen, Jiaming and Liu, Tianqi and Liu, Jialu and Metzler, Donald and others},
  booktitle={Findings Assoc. Comput. Linguistics: NAACL},
  pages={1504--1518},
  year={2024}
}

@inproceedings{zhuang2024setwise,
  title={A setwise approach for effective and highly efficient zero-shot ranking with large language models},
  author={Zhuang, Shengyao and Zhuang, Honglei and Koopman, Bevan and Zuccon, Guido},
  booktitle={Proc. 47th Int. ACM SIGIR Conf.},
  pages={38--47},
  year={2024}
}

@inproceedings{chen2025accelerating,
  title={Accelerating listwise reranking: Reproducing and enhancing {FIRST}},
  author={Chen, Zijian and Pradeep, Ronak and Lin, Jimmy},
  booktitle={Proc. 48th Int. ACM SIGIR Conf.},
  pages={3165--3172},
  year={2025}
}

@article{pradeep2023rankzephyr,
  title={{RankZephyr}: Effective and robust zero-shot listwise reranking is a breeze!},
  author={Pradeep, Ronak and Sharifymoghaddam, Sahel and Lin, Jimmy},
  journal={arXiv:2312.02724},
  year={2023}
}

@inproceedings{xu2024list,
  title={List-aware reranking-truncation joint model for search and retrieval-augmented generation},
  author={Xu, Shicheng and Pang, Liang and Xu, Jun and Shen, Huawei and Cheng, Xueqi},
  booktitle={Proc. ACM Web Conf.},
  pages={1330--1340},
  year={2024}
}

@article{cheng2026enhancing,
  title={Enhancing financial report question-answering: A retrieval-augmented generation system with reranking analysis},
  author={Cheng, Zhiyuan and Lai, Longying and Liu, Yue and Cheng, Kai and Qi, Xiaoxi},
  journal={arXiv:2603.16877},
  year={2026}
}

@inproceedings{thorne2018fever,
  title={{FEVER}: A large-scale dataset for fact extraction and verification},
  author={Thorne, James and Vlachos, Andreas and Christodoulopoulos, Christos and Mittal, Arpit},
  booktitle={Proc. NAACL-HLT},
  pages={809--819},
  year={2018}
}

@inproceedings{marelli2014sick,
  title={A {SICK} cure for the evaluation of compositional distributional semantic models},
  author={Marelli, Marco and Menini, Stefano and Baroni, Marco and Bentivogli, Luisa and Bernardi, Raffaella and Zamparelli, Roberto},
  booktitle={Proc. 9th Int. Conf. Lang. Resources Eval. (LREC)},
  pages={216--223},
  year={2014}
}

@inproceedings{nie2019combining,
  title={Combining fact extraction and verification with neural semantic matching networks},
  author={Nie, Yixin and Chen, Haonan and Bansal, Mohit},
  booktitle={Proc. AAAI Conf. Artif. Intell.},
  volume={33},
  pages={6859--6866},
  year={2019}
}

@article{nguyen2016msmarco,
  title={{MS MARCO}: A human generated machine reading comprehension dataset},
  author={Nguyen, Tri and Rosenberg, Mir and Song, Xia and Gao, Jianfeng and Tiwary, Saurabh and Majumder, Rangan and Deng, Li},
  journal={arXiv:1611.09268},
  year={2016}
}

@inproceedings{craswell2020trec19,
  title={Overview of the {TREC} 2019 deep learning track},
  author={Craswell, Nick and Mitra, Bhaskar and Yilmaz, Emine and Campos, Daniel and Voorhees, Ellen M},
  booktitle={Proc. 28th Text REtrieval Conf. (TREC)},
  year={2020}
}

@inproceedings{craswell2021trec20,
  title={Overview of the {TREC} 2020 deep learning track},
  author={Craswell, Nick and Mitra, Bhaskar and Yilmaz, Emine and Campos, Daniel},
  booktitle={Proc. 29th Text REtrieval Conf. (TREC)},
  year={2021}
}

@inproceedings{xiao2024cpack,
  title={{C-Pack}: Packed resources for general {Chinese} embeddings},
  author={Xiao, Shitao and Liu, Zheng and Zhang, Peitian and Muennighoff, Niklas and Lian, Defu and Nie, Jian-Yun},
  booktitle={Proc. 47th Int. ACM SIGIR Conf.},
  pages={641--649},
  year={2024}
}

@inproceedings{longpre2021entity,
  title={Entity-based knowledge conflicts in question answering},
  author={Longpre, Shayne and Perisetla, Kartik and Chen, Anthony and Ramesh, Nikhil and DuBois, Chris and Singh, Sameer},
  booktitle={Proc. Conf. Empirical Methods Natural Lang. Process. (EMNLP)},
  pages={7052--7063},
  year={2021}
}

@inproceedings{hou2024wikicontradict,
  title={{WikiContradict}: A benchmark for evaluating {LLMs} on real-world knowledge conflicts from {Wikipedia}},
  author={Hou, Yufang and Pascale, Alessandra and Carnerero-Cano, Javier and Tchrakian, Tigran and Marinescu, Radu and Daly, Elizabeth and Padhi, Inkit and Sattigeri, Prasanna},
  booktitle={Adv. Neural Inf. Process. Syst.},
  volume={37},
  year={2024}
}

@inproceedings{hasan2025kag,
  title={{KAG}: A scalable knowledge-augmented generation system for educational content management},
  author={Hasan, Hadi and Ismail, Ali and Mohanna, Ammar and Chehab, Ali},
  booktitle={Proc. 3rd Int. Conf. Foundation Large Lang. Models (FLLM)},
  pages={503--508},
  year={2025}
}

@inproceedings{hasan2026toward,
  title={Toward trustworthy large language model agents in healthcare},
  author={Hasan, Hadi and Salman, Safaa and Abou Dargham, Adam Tai and Mohanna, Ammar and Chehab, Ali},
  booktitle={Proc. 5th Int. Conf. Comput. Netw., Big Data IoT (ICCBI)},
  pages={938--945},
  year={2026}
}

@inproceedings{hasan2026defending,
  title={Defending against adaptive adversarial attacks: A reinforcement learning-based approach to deep neural network robustness},
  author={Hasan, Hadi and Chehab, Ali and Tajeddine, Razane},
  booktitle={Proc. IEEE 5th Int. Multidisciplinary Conf. Eng. Technol. (IMCET)},
  pages={60--66},
  year={2026}
}

@article{hasan2026end,
  title={End-to-end deep learning pipeline for opportunistic vertebral fracture and {Schmorl's} node detection from routine {CT} scans},
  author={Hasan, Hadi and El-Hajj Fuleihan, Ghada and Natout, Mustafa and Mohanna, Ammar and Tajeddine, Razane and Rafeh, Walid and Saleh, Firas and El Alam, Raquelle and Chahine, Reve and Khoury, Nabil and others},
  journal={Frontiers Musculoskeletal Disorders},
  volume={4},
  pages={1718482},
  year={2026}
}

@article{hasan2026comprehensive,
  title={Comprehensive benchmarking of deep learning architectures for lung cancer histopathology},
  author={Hasan, Hadi and Salman, Safaa and Sleem, Lama and Mouawad, Ralph and Chehab, Ali},
  journal={arXiv:2608.15915},
  year={2026}
}

\end{document}